\documentclass[a4paper]{cas-dc}
\usepackage{amssymb}
\usepackage{amsmath,amsfonts}
\usepackage{algorithmic}
\usepackage{algorithm}
\usepackage{array}
\usepackage[caption=false,font=normalsize,labelfont=sf,textfont=sf]{subfig}
\usepackage{textcomp}
\usepackage{stfloats}
\usepackage{url}
\usepackage{verbatim}
\usepackage{graphicx}

\newproof{pf}{Proof}
\newtheorem{theorem}{Theorem}
\newtheorem{lemma}{Lemma}
\usepackage{algorithm}
\usepackage{algorithmic}

\usepackage{newfloat}
\usepackage{listings}
\usepackage{booktabs}
\usepackage{amsmath}
\usepackage{pifont}
\usepackage[numbers,sort&compress]{natbib}
\begin{document}
\shorttitle{ }
\shortauthors{Zongbin Wang et~al.}
\title [mode = title]{Domain Agnostic Conditional Invariant Predictions for Domain Generalization}
\tnotemark[1]
\tnotetext[1]{The work was supported by the National Key Research and Development Program of China under Grant 2022YFA1003800 and 2022ZD0160401, the National Natural Science Foundation of China under the Grant 62001251 and 62125102, and the Beijing-Tianjin-Hebei Basic Research Cooperation Project under the Grant F2021203109.(Corresponding author: Bin Pan.)}

\author[1]{Zongbin Wang}[orcid=0009-0004-5808-147X]
\fnmark[1]

\author[1]{Bin Pan}
\cormark[1]
\fnmark[2]
\ead{panbin@nankai.edu.cn}

\author[2]{Zhenwei Shi}
\fnmark[3]

\cortext[mycorrespondingauthor]{Corresponding author}
\address[1]{School of Statistics and Data Science, Nankai University, Tianjin 300071, PR China}
\address[2]{Image Processing Center, School of Astronautics, Beihang University, Beijing 100191 , PR China}



	\begin{abstract}
    	Domain generalization aims to develop a model that can perform well on unseen target domains by learning from multiple source domains. However, recent-proposed domain generalization models usually rely on domain labels, which may not be available in many real-world scenarios. To address this challenge, we propose a Discriminant Risk Minimization (DRM) theory and the corresponding algorithm to capture the invariant features without domain labels. In DRM theory, we prove that reducing the discrepancy of prediction distribution between overall source domain and any subset of it can contribute to obtaining invariant features. To apply the DRM theory, we develop an algorithm which is composed of Bayesian inference and a new penalty termed as Categorical Discriminant Risk (CDR). In Bayesian inference, we transform the output of the model into a probability distribution to align with our theoretical assumptions. We adopt sliding update approach to approximate the overall prediction distribution of the model, which enables us to obtain CDR penalty. We also indicate the effectiveness of these components in finding invariant features. We evaluate our algorithm against various domain generalization methods on multiple real-world datasets, providing empirical support for our theory.
	\end{abstract}
\maketitle

	\section{Introduction}
    Despite significant progress made by deep learning models in various fields \cite{ref68,ref75,ref81,ref88}, a significant portion of these models can still fail due to data distributions shifting\cite{ref69,ref72,ref74,ref78,ref76}. This limitation hinders their deployment in practical applications. The failure is often attributed to models relying on irrelevant "spurious features" that do not determine the class labels. For instance, a related study\cite{ref54} highlighted a possible failure case where a model classified cows and camels. The background serves as a type of feature that is not directly related to the true labels and can arbitrarily vary across domains. Evidently, the poor generalization performance of the model is attributable to the utilization of all the features in pictures that may be associated with the labels\cite{ref26,ref36,ref74,ref79}.		

	The aforementioned problem can be characterized as domain generalization problem, which can be defined as follows: training a model on multiple source domains with the objective of achieving stable and reliable predictions on new and unseen target domains\cite{ref68}. Compared to the related domain adaptation problem\cite{ref64,ref77,ref71,ref80,ref66}, domain generalization models do not have access to the target domain during training, and this is also difficult to apply domain adaptation theories directly to domain generalization\cite{ref57}, rendering the problem considerably more challenging. Many studies have been conducted in this area, and several types of methods have been proposed, these methods can be broadly categorized into the following groups: \textbf{penalty}: including IRM \cite{ref3}, VREx \cite{ref19} ; \textbf{alignment}: including CORAL \cite{ref8}, MMD \cite{ref56}; \textbf{sample reweighting}: such as GDRO \cite{ref34} ; \textbf{adversarial}: such as DANN \cite{ref12}; additionally, there are also methods to improve the generalization ability of the model by data augmentation \cite{ref41} or integrated model \cite{ref4,ref7}, etc.

	Despite the potential benefits of domain generalization, several challenges hinder its widespread adoption in practice. One key obstacle is that many existing models need group or domain labels, which precisely define the distribution from which the algorithm should be independent or stable \cite{ref65,ref3}. Domain labels are often difficult to obtain or define in real-world scenarios due to various factors, such as high costs, privacy concerns, or lack of domain knowledge\cite{ref9,ref67}. Consider the example of images available on the Internet, where only the category is available, and domain information is unknown. 
	
	To address the issue of unavailable domain labels, we propose a new theory called Discriminant Risk Minimization (DRM), which introduces a new risk metric, Discriminant Risk, to measure the difference between model prediction distributions on the entire source domain and subsets without domain labels. We argue that when a model relies on spurious features for prediction, its prediction distribution may vary greatly with different subsets of source domains, regardless of its performance on the overall source domain. We believe that models relying on spurious features for prediction will have higher Discriminant Risk, while using invariant features will result in a stable prediction distribution and lower Discriminant Risk. Furthermore, we prove an upper bound for Discriminant Risk on the target domain, indicating that reducing Discriminant Risk on the source domain can also help minimize that on the target domain. Our theoretical analysis demonstrates the effectiveness of Discriminant Risk and its potential for generalization.	
	
	
	Based on our theoretical findings, we propose a new algorithm to practically apply our theory. We incorporate Bayesian inference and introduce a simplified penalty called Categorical Discriminant Risk (CDR), resulting in improved generalization ability. By introducing Bayesian inference, the output of the model is interpreted as a probability distribution, which introduces uncertainty to prevent overfitting and improve generalization ability. Moreover, we can utilize  these prediction distributions to approximate the overall prediction distribution of the model, generating CDR penalty to minimize differences between the model prediction distributions on partial and overall data by practice.

	Our contributions include:
		
	\begin{enumerate}
	    \item[1.]We propose a new theory named DRM to improve model generalization ability. In which we introduce a new risk, Discriminant Risk, that quantifies reliance of model on spurious features without requiring domain labels.
	    \item[2.]We provide theoretical upper bound for Discriminant risk. This shows that by reducing Discriminant Risk in the source domain, we can reduce Discriminant Risk in the target domain. The finding suggests that decreasing this risk in the source domain leads to improved prediction stability in target domain.
	    \item[3.]Based on above findings, we develop a new algorithm. We introduce Bayesian inference and a simplified penalty called Categorical Discriminant Risk to correct the reliance on spurious features by practically utilizing Discriminant Risk.

	\end{enumerate}
	\section{Related Work}
  	 In the past few years, there has been a significant focus on domain generalization, due to its applicability in practical scenarios. Below, we present a portion of the relevant literature in this area.

    \textbf{ERM}: The first and classic approach is Empirical Risk Minimization\cite{ref2}, which solely requires to minimize the empirical risk on the source domain. Despite the fact that the target domain may differ from the source domain, ERM typically yields competitive results\cite{ref13}. 
     
     \textbf{Penalty}: Models like IRM\cite{ref3} and VREx\cite{ref19} assume an optimal, invariant feature $z_{inv}$ determines the response variable $Y$, but direct identification from data is challenging. These models use penalty methods to achieve their objectives, such as IRM finding an optimal classifier for all domains and VREx reducing prediction variance across source domains. Despite the strong theoretical guarantees for near-linear models, deep learning models tend to overfit and perform comparably to ERM\cite{ref2,ref6}. To improve IRM in deep models, research has explored Bayes-IRM \cite{ref23} and other methods.     
    
	\textbf{Align and Adversarial Model }: Approaches like DANN\cite{ref12}, CDANN\cite{ref21}, \cite{ref38} aspires to cultivate invariant features by incorporating adversarial modules, enabling discrimination across diverse source domains. Conversely, methods such as MMD\cite{ref56}, CORAL\cite{ref8}, and \cite{ref58,ref59} reshape source domain features into a shared space, fostering their resemblance and fortifying the model generalization capacity. Simultaneously, \cite{ref64,ref67} leverage the VAE method to segregate invariant and spurious features for prediction. Additionally, there is a consideration to utilize alignment for enhancing both fairness and model generalization\cite{ref30}. Despite aligning feature distributions through various losses, researchers have uncovered that the inclusion of spurious invariant features can still deceive the domain discriminator, rendering the structure ineffective\cite{ref26,ref57}. Furthermore, the performance of these models might be constrained by the use of over-parameterized models\cite{ref55}.

    Apart from the above methods, numerous other excellent techniques exist. These include the Mixstyle\cite{ref43} approach, which utilizes style transfer to enhance model performance, and the \cite{ref39} technique for fine-tuning on test set data. Additionally, \cite{ref34,ref24} rely on sample reweighting; Fish\cite{ref37}, Fishr\cite{ref31}, SelfReg\cite{ref18}, SD\cite{ref29} and SAGM\cite{ref62} improve model stability from the perspective of domain gradient update, while \cite{ref61} attempts to treat the loss function for domain generalization from a probabilistic point of view, all of which are also noteworthy.
	\section{Discriminant Risk Minimization}
	\subsection{Problem setting}
	
	\begin{table}
		\centering
		\caption{Mathematical notation }
		\label{math1}    			
		\resizebox{0.45\textwidth}{!}{
			\begin{tabular}{cl}
				\bottomrule
				\specialrule{0em}{1pt}{1pt}
				\hline
				Symbol   & Description         \\ \hline
				\specialrule{0em}{0.5pt}{0.5pt}
				$\mathcal{E}$ & Domain label set \\
				$D$ &  Entire datasets, $D=\bigcup_{e=1}D_e,e\in \mathcal{E}$\\
				$D_s$ & Source domain, $D_s \subset D$   \\
				$D_t$ & Target domain, $D_t \subset D$ \\ 
				$\mathcal{Q}$& Classification function hypothesis space   \\	 
				$\Phi$ & Feature extraction function hypothesis space\\
				$f$ & Total model, $f = q \cdot \phi, \forall q\in \mathcal{Q},\phi \in \Phi$ \\ $R$ & Task loss\\ $\mathcal{Z}$ & Feature space \\ $\mathcal{X}$ & Covariates space \\ $z_{inv}$ & Invariant feature \\$z_{sup}$ & Spurious feature \\$\Sigma(\cdot)$ &Covariance matrix \\$d_{JS}(p_1,p_2)$ & The root of $JS(p_1,p_2)$\\$d_{KL}(p_1,p_2)$& The root of $KL(p1,p2)$\\ \hline
				\specialrule{0em}{1pt}{1pt}
				\bottomrule
		\end{tabular}}
	\end{table}
	Given the generality of distribution changes, it is essential to define them explicitly. The Table \ref{math1} definitions  followed by \cite{ref3,ref13} can be considered.
	
     During training, we only have access to source domain $D_s$. If we want to obtain models that can make good predictions on target domain $D_t$ which is not acceptable during training, in most cases, we can directly use the Empirical Risk Minimization, or ERM to optimize the model: 
    \begin{equation}
    	\phi_0,q_0 \in \mathop{\arg\min}\limits_{\phi\in\Phi,q\in\mathcal{Q}}R_{D_s}(Y,q\cdot\phi(X)) .\label{erm}
    \end{equation}
        However, under our assumptions, the ERM is not ideal because the $D_s$ and $D_t$ do not follow the same distribution. So we make the assumption that there exists a feature space $\mathcal{Z}$ that generates samples $\mathcal{X}$, $g:z\rightarrow x$. We can divide any $z$ into two non-overlapping parts: $z = (z_{inv}, z_{sup})$, the $z_{inv}$ is invariant features, as we assume that the distribution of $z_{inv}$ remains unchanged as domain label $e$ varies: 
        \begin{equation}
        	P(z_{inv}|D_i)=P(z_{inv}|D_j),\forall i,j\in \mathcal{E}.
        \end{equation}
         We also suppose the $z_{inv}$ determines the data label $h:z_{inv}\rightarrow y$, for example , we can identify an animal by its outline, whether it's a cartoon or photograph, so the outline serves as a $z_{inv}$. In contrast, $z_{sup}$ is considered as a spurious feature that does not determine the label but is strongly correlated with it. The distribution of $z_{sup}$ changes drastically with the domain:
         \begin{equation}
         	P(z_{sup}|D_i)\neq P(z_{sup}|D_j), \forall i,j\in \mathcal{E}.
         \end{equation}
       
	Since we need to find a model that can perform well on any target domain $D_t$, based on our assumptions, such a model should not rely on $z_{sup}$ because they may vary acutely causing the model to fail, in other words, we need to find a $\phi_0\in\Phi$ that can satisfy $\phi_0(x_i)=z_{inv}$. So we set our optimization goal:

 \begin{equation}
 	\begin{split}
 	q_0&\in\mathop{\arg\min}\limits_{q\in\mathcal{Q}}R_{D_{s}}(Y,q\cdot\phi_0(X))
    \\&s.t. \phi_0(x) = z_{inv},\forall x \in D_s.
 	\end{split}
   \end{equation}

     \subsection{Discriminant Risk for domain generalization}
      For simplicity, we make: $P_{D_i}(\hat{y}|q,\phi)$ as the prediction distribution made by model $q \cdot \phi$ in dataset $D_i \subset D$.
      We begin with a straightforward hypothesis: In the datasets $D$ within the same class, $z_{sup}$ exhibits higher uncertainty compared to $z_{inv}$. This implies $det(\Sigma(z_{sup})) > det(\Sigma(z_{inv}))$. We find this assumption reasonable since image categorization often requires only a small amount of information, yet the photos may vary significantly in terms of angles, styles, and lighting conditions based on their domain labels.
      
      As discussed in the preceding section, we aim to find a model utilizing $z_{inv}$ for predictions. To assess whether the model employs $z_{inv}$, we turn to \cite{ref63}, which establishes a connection between data uncertainty and model prediction uncertainty: 
      \begin{equation}
      	\Sigma(\hat{y}) \propto \Sigma(z), \forall \hat{y} \sim P_D(\hat{y}|q,\phi).
      \end{equation}
      According to our hypothesis, predictions using $z_{sup}$ will inherently possess greater uncertainty than those using $z_{inv}$. In other words, when randomly selecting $D_k \subset D$ at this stage, the prediction uncertainty with $z_{sup}$ should surpass that with $z_{inv}$ in $D_k$: $det(\Sigma(\hat{y}_{sup})) > det(\Sigma(\hat{y}_{inv}))$. We posit that in randomly sampled data, $z_{sup}$ might signify spurious features among diverse image types, spanning from sketches to composites, resulting in a substantial variance. However, relying solely on $\Sigma(\hat{y})$ or $\Sigma(z)$ for evaluation is suboptimal, as the model might decrease $\Sigma(\hat{y})$ on the source domain through overfitting while simultaneously increasing $\Sigma(\hat{y})$ on the target domain.
      
     
       Inspired by this, we define the following risk to measure the uncertainty of  model predictions: we reduce the task to a binary classification problem and provide the following definition: for any given model $f=q\cdot\phi$, we call:
      \begin{equation}
     \varepsilon_{D_i}^f=\int |P_{D_i}(\hat{y}|q,\phi)-P_{D_s}(\hat{y}|q,\phi)|dx, \label{ds}
     \end{equation}
     the Discriminant Risk of $D_s$ for any set $D_i\subset D_s$. We use this loss to quantify the difference in prediction performance between $D_i$ and the overall source domain $D_s$. 
     
     We represent the model capturing invariant features as $f_{z_{inv}}$ and the one capturing spurious features as $f_{z_{sup}}$. When model prediction accuracies are similar, we can modify the form of $\varepsilon_{D_i}^f = \iint P(\hat{y}|q,\phi,x)|dP_{D_i}-dP_{D_s}|dx$. Under the same $D_i$, the risk value is primarily influenced by the disparity between the locally and globally adopted feature distributions by the model, which is linked to the variance of the features $z$. In binary categories, $\varepsilon_{D_i}^f \propto var(\hat{y})$. Given that $var(\hat{y}_{sup}) > var(\hat{y}_{inv})$, most models will satisfy the inequality $\varepsilon_{D_i}^{f_{z_{sup}}} \geq \varepsilon_{D_i}^{f_{z_{inv}}}, D_i \subset D_s$. This holds true for the target domain as well: $\varepsilon_{D_t}^{f_{z_{sup}}} \geq \varepsilon_{D_t}^{f_{z_{inv}}}, \forall D_t \subset D$. The definition of $\varepsilon_{D_t}^f$ is $\varepsilon_{D_t}^f = \int | P_{D_t}^f(\hat{y}) - P_{D_s}^f(\hat{y}) | dx$, noting that $D_t \not\subset D_s$, but $D_t, D_s \subset D$. Our assumptions about $z$ still hold, where models using $z_{sup}$ exhibit greater uncertainty in $D_t$ than models using $z_{inv}$ for predictions.
     
     As the decrease in $\varepsilon_{D_t}^f$ occurs, the model places greater reliance on invariant features for predictions. This is attributed to the alignment of predictive distributions in both the source and target domains, indicating the utilization of similar features in prediction. According to our assumptions, this alignment implies the use of invariant features, ultimately enhancing predictions. Consequently, mitigating Discriminant Risk holds the potential to enhance model accuracy in target domains.
     
     Given the adverse effects of incorporating the spurious feature $z_{sup}$ in the model, our primary goal is to enhance predictive performance on the target domain by minimizing both ERM and $\varepsilon_{D_t}^f$. Since information from the target domain is unavailable during training, we must explore alternative methods to decrease $\varepsilon_{D_t}^f$. In pursuit of this objective, we propose a theorem to establish the relationship between $\varepsilon_{D_t}^f$ and $\varepsilon_{D_i}^f,D_i \subset D_s.$
     
     To introduce our theorem, we first present two lemmas and their proofs.
     
     For convenience, note that $d_{js}(P_i,P_j)=\sqrt{JS(P_i,P_j)}$, which is the square root of the JS divergence of any two distributions, where the JS divergence is the Jensen-Shannon divergence. First we introduce the KL divergence, $KL(p,q)=\int p ln(\frac{p}{q})dx$, and JS divergence is defined based on this: 
      \begin{equation}
     JS(p,q)=\frac{1}{2}(KL(p,p')+KL(q,p')),p'=\frac{p+q}{2}.
      \end{equation}
      
      \begin{lemma}
      \item For $\forall D_i,D_j\subset D_s$, for simplicity, let $P_{D_s^i}^f(\hat{y})$ represent the $P_{D_s^i}(\hat{y}|q,\phi), f=q\cdot\phi$, we have:
      \begin{equation}
      	\int|P_{D_i}^f(\hat{y})-P_{D_j}^f(\hat{y})|dx\leq\sqrt{2}d_{JS}(P_{D_j}^f(\hat{y}),P_{D_i}^f(\hat{y})).
      \end{equation}
    \end{lemma}
      \begin{pf}

     First we define a loss in $D_{ij}=D_i\cup D_j$, $d_{KL}(p,q) = \sqrt{KL(p,q)}$, we have:
      \begin{equation}
      	\int|P_{D_i}^f(\hat{y})-P_{D_{ij}}^f(\hat{y})|dx\\
      	\leq\frac{1}{\sqrt{2}}d_{KL}(P_{D_i}^f(\hat{y}),P_{D_{ij}}^f(\hat{y})).
      \end{equation}
      The above inequality is a direct application of Pinsker inequality. Similarly we have:
      \begin{equation}
      	\int|P_{D_j}^f(\hat{y})-P_{D_{ij}}^f(\hat{y})|dx\\
      	\leq\frac{1}{\sqrt{2}}d_{KL}(P_{D_j}^f(\hat{y}),P_{D_{ij}}^f(\hat{y})),
      \end{equation}
      then we can prove:
      \begin{align}
      		&\int|P_{D_j}^f(\hat{y})-P_{D_i}^f(\hat{y})|dx\\&=\int|P_{D_j}^f(\hat{y})-P_{D_{ij}}^f(\hat{y})+P_{D_{ij}}^f(\hat{y})-P_{D_i}^f(\hat{y})|dx\\&
      		\leq \int|P_{D_j}^f(\hat{y})-P_{D_{ij}}^f(\hat{y})|+|P_{D_{ij}}^f(\hat{y})-P_{D_i}^f(\hat{y})|dx\label{tr}\\&
      		\leq \frac{1}{\sqrt{2}}(d_{KL}(P_{D_i}^f(\hat{y}),P_{D_{ij}}^f(\hat{y}))+ d_{KL}(P_{D_j}^f(\hat{y}),P_{D_{ij}}^f(\hat{y})))\\&
      		\leq \sqrt{2} d_{js}(P_{D_i}^f(\hat{y}),P_{D_j}^f(\hat{y})).\label{ca} 
      \end{align}
        \end{pf}
      The inequality (\ref{tr}) is held through the Triangle Inequality, while (\ref{ca}) is established using the Cauchy-Schwarz Inequality.
      \begin{lemma}

     \item According to Lemma 1, for $\forall D_i,D_j\subset D_s$, giving the model $f$, we are able to infer:
      \begin{equation}
      	\centering
      	\varepsilon_{D_i}^f\leq\varepsilon_{D_j}^f+2d_{JS}(P_{D_j}^f(\hat{y}),P_{D_i}^f(\hat{y})).
      \end{equation}
       \end{lemma}
       \begin{pf}

        Using Lemma 1 and for any $\varepsilon_{D_i}^f$ and $f = q \cdot \phi$, we have:
      \begin{align}
      		\varepsilon_{D_i}^f&=\int|P_{D_i}^f(\hat{y})-P_{D_s}^f(\hat{y})|dx\\&=\int|P_{D_i}^f(\hat{y})-P_{D_j}^f(\hat{y})+P_{D_j}^f(\hat{y})-P_{D_s}^f(\hat{y})|\\&
      		\leq \int|P_{D_i}^f(\hat{y})-P_{D_j}^f(\hat{y})|+|P_{D_j}^f(\hat{y})-P_{D_s}^f(\hat{y})|\\&
      		\leq\varepsilon_{D_j}^f+\sqrt{2}d_{JS}(P_{D_i}^f(\hat{y}),P_{D_j}^f(\hat{y})).
      \end{align}
         \end{pf}
      
      We can now state our Theorem 1:
     
     \begin{theorem}(Upper bound) Based on our assumptions about the problem, we can randomly split the source data into multiple disjoint subsets:
     $D_s^i,D_s^j\subset D_s,D_s^i\cap D_s^j=\emptyset,\forall i,j\in(1,2,\ldots,N),\bigcup_{i=1}^{N}{D_s^i=D_s}$, we assume that:
     \begin{equation}
     	\mathop{\max}\limits_i{d_{JS}(P_{D_s^i}^f(\hat{y}),P_{D_s}^f(\hat{y}))}\geq d_{JS}(P_{D_s^u}^f(\hat{y}),P_{D_s^v}^f(\hat{y})), 
     \end{equation}
     \begin{equation}
        \forall u,v\in(1,2,3,\ldots,N), \nonumber
     \end{equation}
      we can prove:
    \begin{equation}
    	\begin{split}
    	\varepsilon_{D_t}^f\leq \frac{1}{N}\sum_{i=1}^N\varepsilon_{D_s^i}^f&+\mathop{\min}\limits_{i}{\sqrt{2}}d_{JS}(P_{D_s^i}^f(\hat{y}),P_{D_t}^f(\hat{y}))\\&+\mathop{\max}\limits_id_{JS}(P_{D_s^i}^f(\hat{y}),P_{D_s}^f(\hat{y})).\label{the}
    	 \end{split}
    \end{equation}
    \end{theorem}
     \begin{pf}

 First we consider $\varepsilon_{D_t}^f=\int| P_{D_t}^f(\hat{y})-P_{D_s}^f(\hat{y})|dx$ and note that if $D_i,D_j \not\subset D_t$, Lemma 1 and 2 still holds, so we can replace $D_i,D_j$ with $D_s^i$ and $D_t$, we have:
    \begin{equation}
    	\varepsilon_{D_t}^f\leq\varepsilon_{D_s^i}^f+d_{JS}(P_{D_s^i}^f(\hat{y}),P_{D_t}^f(\hat{y})),
    \end{equation}
    and we suppose 
   	$\mathop{\bigcup}\limits_{i=1}^ND_s^i=D_s$
   , we can average it on all $D_s^i$:
    \begin{equation}
    	\varepsilon_{D_t}^f \leq\frac{1}{N}\sum_{i=1}^N\varepsilon_{D_s^i}^f+\frac{\sqrt{2}}{N}\sum_{i=1}^Nd_{JS}(P_{D_s^i}^f(\hat{y}),P_{D_s}^f(\hat{y})).
    \end{equation}
    Suppose there exists a $D_s^*$ nearest to the $D_t$, since JS divergence is a distance metric, then using triangle inequality, we have:
    \begin{align}
    		\varepsilon_{D_t}^f&\leq\frac{1}{N}\sum_{i=1}^N\varepsilon_{D_s^i}^f+\frac{\sqrt{2}}{N}\sum_{i=1}^Nd_{JS}(P_{D_s^i}^f(\hat{y}),P_{D_s}^f(\hat{y}))\\
    		&\leq\frac{1}{N}\sum_{i=1}^N\varepsilon_{D_s^i}^f+{\sqrt{2}}d_{JS}(P_{D_*}^f(\hat{y}),P_{D_s}^f(\hat{y}))\nonumber\\&+\frac{\sqrt{2}}{N}\sum_{i=1}^Nd_{JS}(P_{D_s^i}^f(\hat{y}),P_{D_*}^f(\hat{y})).
    \end{align}
    Combining the assumptions that:
    \begin{equation}
    	\mathop{\max}\limits_i{d_{JS}(P_{D_s^i}^f(\hat{y}),P_{D_s}^f(\hat{y}))}\geq d_{JS}(P_{D_s^i}^f(\hat{y}),P_{D_s^j}^f(\hat{y})),
    \end{equation}
    \begin{equation}
    	\forall i,j\in(1,2,3,\ldots,n), \nonumber
    \end{equation}
     we can easily deduce that:
    \begin{align}
    		&\frac{1}{N}\sum_{i=1}^N\varepsilon_{D_s^i}^f+{\sqrt{2}}d_{JS}(P_{D_*}^f(\hat{y}),P_{D_s}^f(\hat{y}))\nonumber\\&+\frac{\sqrt{2}}{N}\sum_{i=1}^Nd_{JS}(P_{D_s^i}^f(\hat{y}),P_{D_*}^f(\hat{y}))\\&\leq \frac{1}{N}\sum_{i=1}^N\varepsilon_{D_s^i}^f+\mathop{\min}\limits_{i}{\sqrt{2}}d_{JS}(P_{D_s^i}^f(\hat{y}),P_{D_s}^f(\hat{y}))\nonumber\\&+\mathop{\max}\limits_id_{JS}(P_{D_s^i}^f(\hat{y}),P_{D_t}^f(\hat{y})),
    \end{align}
     therefore:
     \begin{align}
     		\varepsilon_{D_t}^f\leq \frac{1}{N}\sum_{i=1}^N\varepsilon_{D_s^i}^f&+\mathop{\min}\limits_{i}{\sqrt{2}}d_{JS}(P_{D_s^i}^f(\hat{y}),P_{D_t}^f(\hat{y}))\nonumber\\&+\mathop{\max}\limits_id_{JS}(P_{D_s^i}^f(\hat{y}),P_{D_s}^f(\hat{y})).
     \end{align}
          \end{pf}
     Please note that a similar conclusion applies to multidimensional situations. Here we do not impose any restriction on $D_s^i$, i.e., we do not need $D_s^i$ to be a certain dataset with domain labels; the above theorem will hold under any combination of $\{D_s^i\}_{i=1}^N$. Theorem 1 tells us that the upper bound of Discriminant Risk in the target domain is determined by three terms, of which the second term is impractical as it pertains to an unknown target domain. Meanwhile, the first term refers to the mean value of Discriminant Risk of each partial set, and the third term corresponds to the square root of Jensen-Shannon divergence between the prediction distribution of partial and the overall data. Therefore, we show that the defined loss is indeed a measure of whether the model is finding invariant features, and thus has the ability to improve the generalization, we can see it is essential to decrease the values of the first and third terms. 
     
     We observe that both SCA\cite{ref58} and G2DM\cite{ref59} propose similar theories. SCA aims to enhance the model generalization ability by mapping features to a dedicated kernel space while reducing intraclass and inter-domain distances. Similarly, G2DM aims to stabilize the model predictive ability by decreasing its prediction distribution distance. However, both methods are theoretically derived under the assumption of known domain labels, limiting their application in our scenario.
     
      The theorem also indicates that reducing $\varepsilon_{D_t}^f$ is achievable by lowering $\varepsilon_{D_i}^f$ for $D_i \subset D_s$, thereby prompting the model to discover invariant features.
       Specifically, both the first and third terms of (\ref{the}) can be optimized through a reduction in JS divergence between the partial data sets and overall data predictions in each class. Thus, we formulate the following DRM optimization objective:
     \begin{equation}
        \mathop{\arg\min}\limits_{q,\phi}R_{D_s}(Y,q\cdot\phi(X)) \nonumber
             \end{equation}
             \begin{equation}
     	s.t. \quad q,\phi\in \mathop{\arg\min}\limits_{q,\phi}JS(P_{D_s,Y_j}(\hat{y}|q,\phi),P_{D_s^i,Y_j}(\hat{y}|q,\phi)),  \tag{DRM}
     	     \end{equation}
     	     \begin{equation}
     	\forall D_s^i\subset D_s, i\in(1,2,...,N),j \in (1,2,..,c).   \nonumber
     \end{equation}
      Here $n$ represents number of sample batches, $c$ stands for number of categories, we want the model not only to be able to classify correctly, but also to have a prediction distribution that does not change drastically as the $D_s^i$ changes for each class. This can only be accomplished if the model does find invariant features, which will reduce the Discriminant Risk.
     \subsection{Empirical experiment} 
\begin{figure*}
     	\centering
     	\includegraphics[width=1\textwidth]{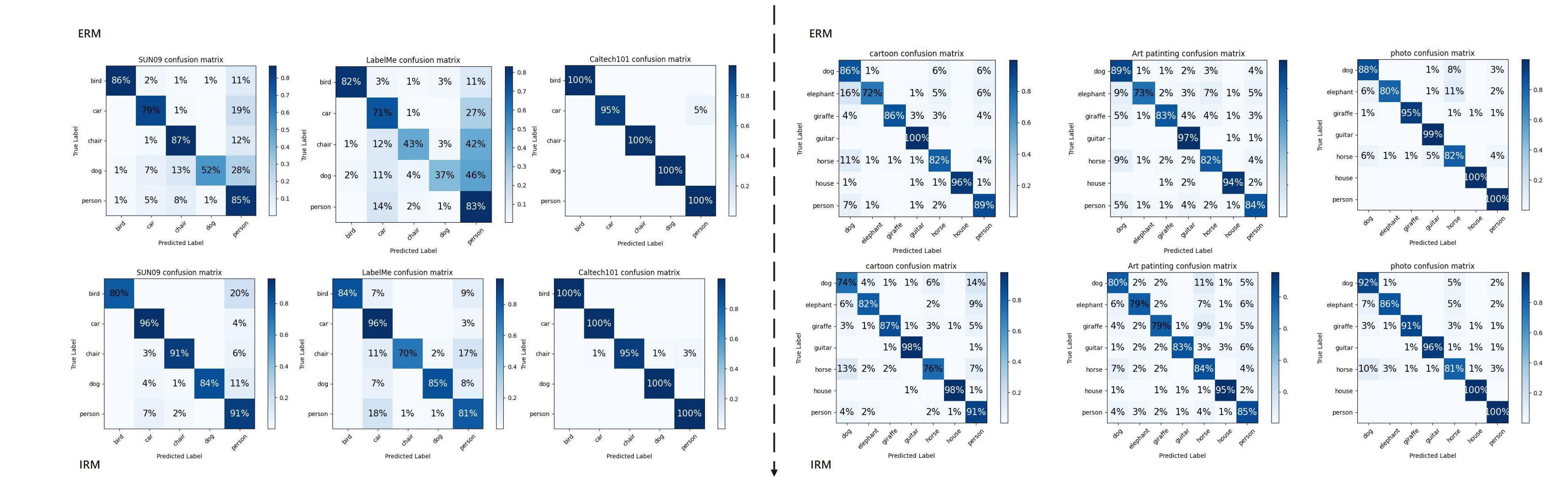}
     	\caption{We performed the aforementioned experiments on resnet-18 using default parameters. The confusion matrix plots on each source domain after 5000 batches on VLCS are displayed on the left side of the image. The number of images in each category is close to one another in each domain. It is apparent that the model exhibits considerable disparities in the confusion matrix for each source domain, even after prolonged training. On the right side, the confusion matrix plots for the same model is trained after 300 batches on PACS. We can observe that the model prediction distribution also exhibits substantial variation among domains due to the dependence on spurious features in the early training stage.}
     	\label{confuxion}
     \end{figure*}

      To examine our hypothesis positing that models lacking invariant features for prediction would manifest heightened uncertainty in predictions, a straightforward approach is to scrutinize distribution disparities within partial data. We leverage classical models for domain generalization to illustrate prediction distributions across various training source domains. Traditional deep learning models, however, are constrained to furnishing a confidence score for classification predictions rather than a probability. Despite this limitation, each model prediction can be construed as a sample from its prediction distribution. By computing the confusion matrix, we can approximate distribution. Each row in the confusion matrix denotes $P_{D_s^i,Y}(\hat{y}|q,\phi)$ for the model prediction within each class. Figure \ref{confuxion} delineates the confusion matrices of ERM and IRM following 5000 training batches on VLCS\cite{ref46} and 300 batches on PACS\cite{ref47}.
 
     ERM is a well-known method that does not use domain labels during training. However, the confusion matrix of the model for each domain indicates significant differences, implying the possibility of capturing spurious features. Meanwhile, IRM exhibits similar deviations for source domains, indicating a failure to capture invariant features. During the early training stage of PACS, we observe significant differences in the confusion matrices among domains. Although the gap between the confusion matrices of the source domains decreases as training progresses, the target domain accuracy does not improve significantly, and sometimes even worsens. 
     
     Our experiments demonstrate that when the model does not use invariant features, the prediction distributions of some of its data will be significantly different from  one another, indicating higher uncertainty for $P_{D_s^i}(\hat{y}|q,\phi)$ . 
     
     \section{Algorithms for Discriminant Risk Minimization}
     \begin{algorithm}[h]
     	\caption{DRM}
     	\label{alg:AOA}
     	\renewcommand{\algorithmicrequire}{\textbf{Input:}}
     	\renewcommand{\algorithmicensure}{\textbf{Output:}}
     	\begin{algorithmic}[1]
     		\REQUIRE Feature extractor $\phi_0$, bayesian linear classifier $b_0$, mixed training data $D_s=\bigcup_{i=1}^ND_i$, distinct matirx $M^0$, update count $c$, training time $C$, scale parameter $\alpha$, batch size $S$, sample times $T$. 
     		\ENSURE learned $\phi_C$, classfier $b_C$    
     		
     		\STATE Initalize $M^0$ = $diag(1,...,1)$, $c = 0$, random initalize $b=(\mu_0,\Sigma_0)$
     		\WHILE{$c\leq C$}
     		\STATE From $D_s$ randomly select $(X,Y)=\{x_i,y_i\}_{i=1}^{S}$ 
     		\STATE Fixed $\phi_c$
     		\FOR{$t \text{ from } 1 \text{ to } T$}
     		\STATE $\hat{b}_c^t = \mu_c + \epsilon_t\Sigma_c,\epsilon_t\sim N(0,I)$
     		\STATE $Y_{pred}^t = \hat{b}_c^t(\phi_c(X))$
     		\ENDFOR
     		\STATE Obtain ELBO by (\ref{vi}) to update $b_c = (\mu_c,\Sigma_c) \rightarrow b_{c+1}=(\mu_{c+1},\Sigma_{c+1})$ 
     		\STATE Fixed $b_{c+1}$
     		\FOR{$t \text{ from } 1 \text{ to } T$}
     		\STATE $\hat{b}_{c+1}^t = \mu_{c+1} + \epsilon_t\Sigma_{c+1},\epsilon_t\sim N(0,I)$
     		\STATE $Y_{pred}^t  = \hat{b}_{c+1}^t(\phi_c(X)) $, update $M^{c-1}\rightarrow M^{c}$ by (\ref{mt}) 
     		\STATE Obtain $loss_1^t,loss_2^t$ by ERM (\ref{erm}), CDR (\ref{cdr})
     		\ENDFOR
     		\STATE $loss = \frac{1}{T}(\sum^T_{t=1}loss_1^t + \alpha loss_2^t)$
     		\STATE Using $loss$ to update $\phi_c \rightarrow \phi_{c+1}$ 
     		\STATE $c = c +1$ 
     		\ENDWHILE
     	\end{algorithmic}
     \end{algorithm}

     It should be noted that the aforementioned optimization objective faces practical challenges. Firstly, for a general classification neural network, the output is deterministic \cite{ref2,ref3}, which can only be interpreted as the confidence of the classification\cite{ref14} instead of prediction distributions. Although we can use the confusion matrix as a reasonable approximation, generating it requires traversing the entire dataset at each update step which is time-consuming. 
     To address these issues, we make two modifications to the model. First, we replace the last layer $q\in \mathcal{Q}$ of the model with a Bayesian linear layer $b \in \mathcal{B}$ same in the\cite{ref25,ref23}. Second, we introduce an additional matrix denoted as the Discriminant matrix $M$ to replace the confusion matrix, which is challenging to optimize. The Discriminant matrix is compared to the output of each batch of the model, resulting in a simplified loss, named CDR.
     
     Thus, at this stage, we have transformed the optimization objective into:
     \begin{equation}
     	\begin{split}
     		\mathop{\arg\min}\limits_{\phi\in\Phi,q(b)\in\mathcal{F} }& E_{q(b)}(R_{D_s}(Y,b\cdot\phi(X)))\\&+ \alpha \sum_Y\sum_j JS(P_{D_s^j,Y}(\hat{y}|b,\phi),m_{Y}),\\& \forall \, Y \in (1,2,...,c), j \in (1,2,...,N), \label{loss}
     	\end{split}
     \
    \end{equation}
      $q(b)$ is an estimate of the posterior distribution $p(b|D_{s})$, which can be seen as a measure of likelihood. On the other hand, the second term in  (\ref{loss}) refers to the CDR penalty, which is based on the introduced matrix $M$. Each row of $M$ represents overall data prediction $m_i = P_{D_s,Y}(\hat{y}|b,\phi),M=(m_1,m_{2},...m_{c})$ for one category. Additionally, the hyperparameter $\alpha$ is used to control the weight of the CDR to balance between improving the training accuracy and promoting the adoption of invariant features. Next, we will introduce these modifications separately in detail.
    
    \paragraph{Variational Inference}
      \begin{figure*}[h]
     	\centering
     	\includegraphics[width=.7\textwidth]{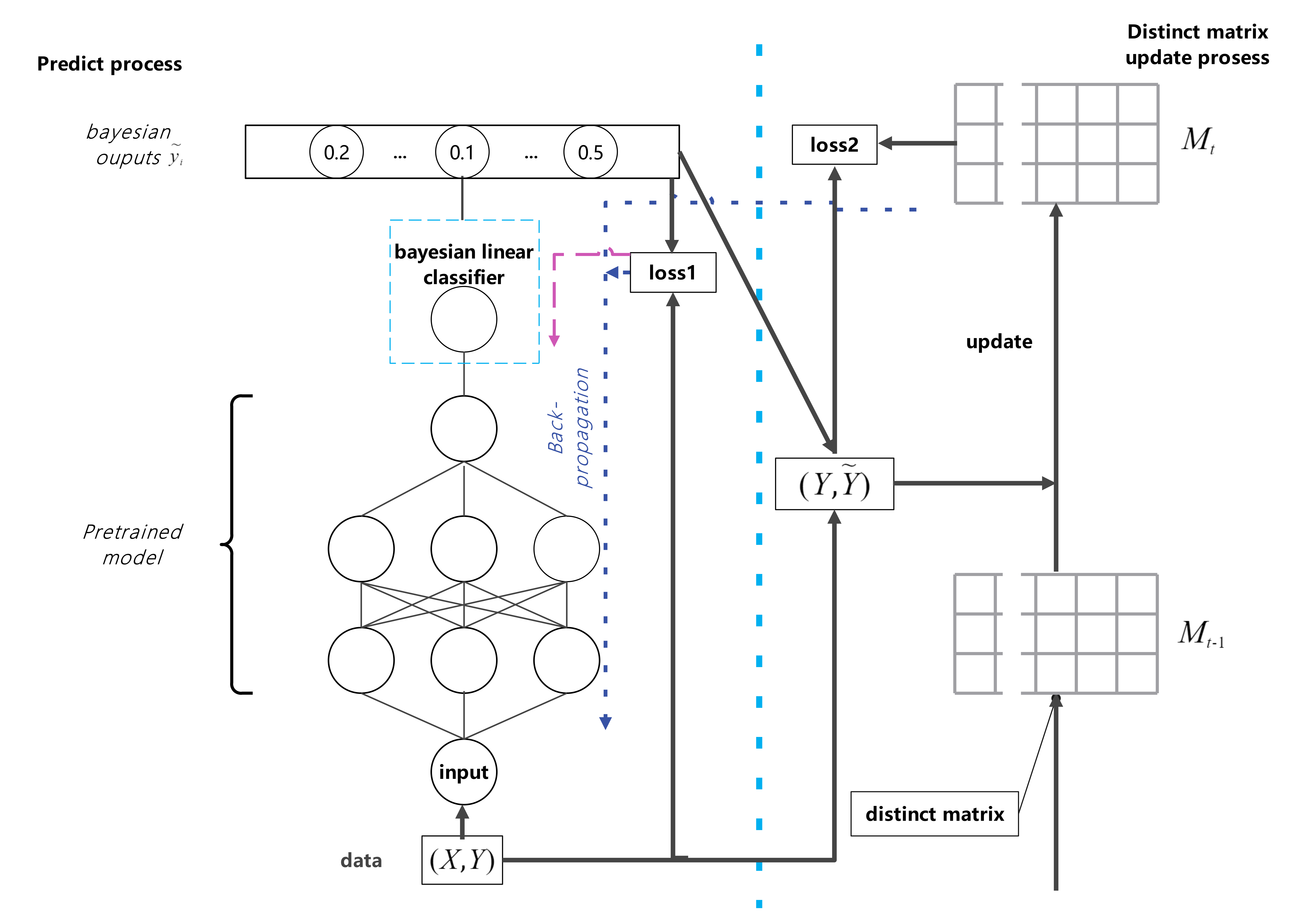}
     	\caption{Our model consists of two distinct parts for updating, which are separated by a blue dashed line. The left-hand side is the prediction process, where VI approximates the posterior distribution $p(b|\mathcal{D}_{s})$ using $q(b)$ to update the Bayesian layer, as shown by the purple dashed backpropagation arrow. The right-hand side is the Discriminant matrix update process, where the Discriminant matrix is used to compare with the model prediction to obtain the CDR. Finally, the feature extractor is updated together with the CDR and likelihood loss.}
     	\label{alg}
     \end{figure*} 
    In Bayes Inference, we want to infer the posterior distribution of the model: $p(b|D_{s}) \propto p(D_s|b)q_0(b)$, however, for an arbitrary $p(b|D_{s})$, it is very difficult to obtain analytic representation, so we use Variational Inference (VI) to approximate $p(b|D_{s})$ using $q(b)$\cite{ref25}, Given a family of distributions $\mathcal{F}$, we use ELBO lower bound to give an optimal $q(b)\in\mathcal{F}$:
    \begin{equation}
    	q(b)=\mathop{\arg\min}\limits_{q'(b)\in\mathcal{F}}E_{q'(b)}(P(D_{s}| \phi,b))-KL(q_0(b), q'(b)) .\label{vi}
    \end{equation}    
    
   We optimize the training data likelihood using the first term and minimize the difference between the approximation distribution and the prior distribution using the second term. For our VI method, we adopt the commonly used approach, selecting a Gaussian distribution $\mathcal{F}=\mathcal{G}=\{N(\mu,\Sigma)|\mu=(h_1,h_2,\ldots,h_n),\Sigma=diag(\sigma_1,...,\sigma_n)\}$, where $n$ is the number of parameters in network last layer. We also assume the prior distribution follows a Gaussian distribution: $q_0(b)=N(0,\sigma I)$. With these settings, we derive the approximate distribution as $q(b)=N(\hat{\mu},\hat{\Sigma})$. 
    
   Furthermore, the Bayesian linear layer provides an added benefit: for the output result $\hat{y}=b\cdot z$, the value of $z$ is deterministic and obtained from the feature extractor output. This means that $\hat{y}$ follows a specific distribution $\hat{y}\sim P(\hat{y}|b)$, resolving the challenge of the output lacking a probability distribution interpretation. Simultaneously, the perturbations in Bayesian layer  significantly enhance the resistance of the model to overfitting. These domain-agnostic techniques substantially improve the model ability to generalize.
    
    \paragraph{Reparameterization} Additionally, the structure of neural networks requires deterministic values for gradient backpropagation, rather than probabilities. It is challenging to determine the analytic solution of the likelihood in the ELBO. Therefore, we use the standard reparameterization method in VI to overcome these challenges: $\hat{b}^t=\mu_b+\epsilon_t \Sigma_b,\epsilon_t \sim N(0,I)$. Thus, by setting the number of resampling to $T$, we change the original likelihood into practical forms: $\frac{1}{T}\sum_{t=1}^{T}\sum_{i=1}^{N}-ln(P(D_s^i|\hat{b}^ t,\phi))$; thus making it practically applicable. Additionally, reparameterization has the added advantage of making the Discriminant matrix applicable, which will be explained in detail in the next subsection.
    
    \paragraph{Categorical Discriminant Risk} We let $P(\hat{y}|b,\phi)$ denote $P_{D_s}(\hat{y}|b,\phi)$, the prediction distribution in $D_s$, which can be expanded as:
    \begin{equation}
     P(\hat{y}|b,\phi)=\iint P_{D_s^i}(\hat{y}|x,b,\phi)p(D_s^i|D_s)dD_s^idx, \label{raw}
    \end{equation}
    where $p(D_s^i|D_s)$ denotes the distribution of partial data in $D_s$. Suppose we have two data points belonging to the same class, $P_{Y}(y_i|x_i,b,\phi)$ and $P_{Y}(y_j|x_j,b,\phi)$, since deep learning networks are typically trained with batches, and in combination with our theoretical $D_s^i$ distribution form, we can group the data points of the same class in each batch as one $D_s^i$
    subset. This approach naturally aligns with our theoretical assumptions, with the probability of each $D_s^i$ occurrence being equal. Thus, suppose we have $K$ separate subset  $D_s^i$, we can consider the following equation:
    \begin{equation}
    P_{Y}(\hat{y}|b,\phi)=\frac{1}{K}\sum_{i=1}^NP_{D_s^i,Y}(\hat{y}|b,\phi),\label{wa}
     \end{equation}
    we approximate the probability $P_{D_s^i,Y}(\hat{y}|b,\phi)$ by averaging over samples: $\frac{1}{n_k}\sum_{j=1}^{n_k}P_{D_s^i,Y}(\hat{y_j}|x_j,b,\phi),(x_j,y_j)\in D_s^i$, enabling us to estimate (\ref{raw}) using (\ref{wa}). But it is challenging to get enough samples in a single batch for accurate estimation within the optimization process. To address this, we leverage the multiple sampling ability of Bayesian classification layer, followed by a softmax function. This lets us treat the output as a categorical distribution, approximating the per-sample prediction distribution as:
    \begin{equation}
    	P_{D_s^i,Y}(\hat{y_j}|x_j,b,\phi) \approx \frac{1}{T}\sum_{t=1}^{T}softmax(b^t \cdot \phi(x_j)) , x_j\in D_s^i,
    \end{equation}
    for simplicity, we denote it as $ y_{pred,f,Y,D_s^i}^j$ . 
    Yet, obtaining the complete $P(\hat{y}|b,\phi)$ after each update is tricky due to gradient updates. To tackle this, we introduce a sliding update matrix, called the Discriminant matrix $M^t=(m_1^t,...,m_c^t)$, of size $c\times c$ during training. The $M^t$ approximates $M = P(\hat{y}|b,\phi)$ over time, with $t$ as the update count and $c$ as the category count. We update the Discriminant matrix using the following equation:
    \begin{equation}
    			m_Y^t=(1-\beta)m_Y^{t-1}+\beta\overline{y}_{pred,f,Y}^t, \label{mt}
    			\end{equation}
    			
    \begin{equation}
    			 \overline{y}_{pred,f,Y}^t=\frac{1}{n_{Y}}\sum_{j=1}^{n_{Y}} y_{pred,f,Y,D_s^t}^j,
       \end{equation}
    where $\overline{y}_{pred,f,Y}^t$ denotes the average prediction for the $Y$th category with $n_{Y}$ instances in $t$th batch $D_s^t \subset D_s$, so it is an approximation for $P_{D_s^t,Y}(\hat{y}|b,\phi)$, $\beta$ is the hyperparameter for sliding update.  Now we change Discriminant Risk into CDR: 
    \begin{equation}
    \hat{\varepsilon}_{D_s^t}^{f}=JS(m_Y^t,\overline{y}^t_{pred,f,Y}), \label{cdr}
    \end{equation} 
     The (\ref{cdr}) represents the CDR penalty used during the $t$th iteration of the gradient update. Consequently, we can optimize the feature extractor $\phi$ using both CDR and ERM. Furthermore, we present three theorems to demonstrate the favorable properties of our structure:
    \begin{theorem}

    (Loss convergence) Assuming that the feature extractor captures $z_{inv}$, as the sample size $N$ tends to infinity, we have:

    \begin{equation}
    	\mathop{\lim}\limits_{t\rightarrow\infty}{JS(m^t_Y,\overline{y}_{pred,f,Y}^t)}{\rightarrow0}.
    \end{equation}
        \end{theorem}  
     \begin{pf}

      When the model captures the invariant features $z_{inv}$, fixed the label $Y$, we have: $\forall (x_i,x_j)\in D_s,P_Y(\hat{y_i}|b,\phi,x_i)=P_Y(\hat{y_j}|b,\phi,x_j)$, since $m_Y^t=(1-\alpha)m_Y^{t-1}+\alpha\overline{y}_{pred,f,Y}^t$, we can expand the formula:
    \begin{align}
    	&m^t_Y = (1-\alpha)^{t-1}m^1_Y+...+\alpha \overline{y}_{pred,f,Y}^{t}\\&
    	m^{t-1}_Y = (1-\alpha)^{t-2}m^1_Y++...+\alpha \overline{y}_{pred,f,Y}^{t-1},
    \end{align}
    $m^Y_1$ is a constant, and we can bring the formula into itself:
    \begin{align}
    		m^t_Y&=(1-\alpha)^{t-1}m^1_Y+(1-\alpha)^{t-2}\alpha \overline{y}_{pred,f,Y}^{2}\\&+(1-\alpha)^{t-3}\alpha \overline{y}_{pred,f,Y}^{3}+...+\alpha \overline{y}_{pred,f,Y}^{t},
    \end{align}
    Suppose the model finds invariant features after $k$ steps, then we have:
    \begin{align}
    		m^t_Y&=(1-\alpha)^{t-1}m^1_Y+(1-\alpha)^{t-2}\alpha \overline{y}_{pred,f,Y}^{2}\\&+(1-\alpha)^{t-3}\alpha \overline{y}_{pred,f,Y}^{3}+...\nonumber\\&+\alpha((1-\alpha)^{t-k}\nonumber\\&+(1-\alpha)^{t-k-1}+...+1)\overline{y}_{pred,f,Y}^{t}\\&=(1-\alpha)^{t-1}m^1_Y+(1-\alpha)^{t-2}\alpha \overline{y}_{pred,f,Y}^{2}\nonumber\\&+(1-\alpha)^{t-3}\alpha \overline{y}_{pred,f,Y}^{3}+...\nonumber\\&+\alpha\frac{1-(1-\alpha)^{t-k}}{\alpha}\overline{y}_{pred,f,Y}^{t}.
    \end{align}
    When the model continues to update, based upon the formula, we can deduce $\overline{y}_{pred,f,Y}^{t}\rightarrow P_f(\hat{y}) $ as $t\rightarrow\infty$, so $JS(m_Y^t,\overline{y}_{pred,f,Y})\rightarrow 0$ holds.
         \end{pf}
    Theorem 2 states that a model capable of capturing invariant features will lead the penalty term to tend towards zero. This results in the model only considering the empirical risk. Conversely, if $\varepsilon_{D_s^t}^f$ is present, it will encourage the model to discard spurious features.
    \begin{theorem}

    (Estimated variance) As above assuming that the model uses $z_{inv}$ for prediction, when the sample size tends to infinity, let the empirical distribution of a single point $x_i \in D_s$ be $\widetilde{P}(\hat{y}) =\sum_{i=1} ^K \frac{I(Y\leq y_{pred}^i)}{K}$, we have: 
    \begin{equation}
    	\mathop{\lim}\limits_{t\rightarrow\infty}var(m^t_Y)\rightarrow var(\widetilde{P}_Y(\hat{y})).
    \end{equation}
        \end{theorem}
      \begin{pf}

      Note that: $\widetilde{P}_Y(\hat{y})=\frac{1}{K}\sum_{i=1}^{K}I(Y_1\leq y_1^{i},…,Y_n\leq y_{n}^{i})$, this is an unbiased estimate and the variance of the estimate is $\frac{P_Y(\hat{y})(1-P_Y(\hat{y}))}{K}$, and we further use this estimation to update the Discriminant matrix. Noting that due to the randomness of the sampling, we assume that each sample is independent of each other, therefore:
    \begin{align}
    		var(m^t_Y)=&((1-\alpha)^{t-2}\alpha)^2var(\widetilde{P}_Y(\hat{y}))\nonumber\\&+((1-\alpha)^{t-3}\alpha)^2var(\widetilde{P}_Y(\hat{y}))\nonumber\\&+...+\alpha^2var(\widetilde{P}_Y(\hat{y})),
    \end{align}
       \end{pf}
    Assuming the model undergoes sufficient updates, we can deduce from the above equation that the prediction variance is primarily influenced by the recent updates. By adjusting $\alpha$, we can determine the impact of previous predictions on the matrix at the current moment. If we assume the model utilizes invariant features for prediction, the equation implies $var(m^t_Y) \rightarrow var(\widetilde{P}_Y(\hat{y}))$. The variance of the empirical distribution approaches that of the true distribution, and at this point, the resampling count $K$ becomes the sole factor influencing results. Multiple samplings can yield stable estimation outcomes.

     This theorem states that if a model uses invariant features to make predictions, then the estimation variance of $m^t_Y$ is similar to the empirical estimate of the same class. And $var(\widetilde{P}_Y(\hat{y}))$ can be expressed as $\frac{c}{K}$, with $c$ being a constant and $K$ being the number of resampling times. Thus, increasing the resampling times helps to reduce uncertainty caused by the sampling.
     \begin{theorem}

   (Sufficient condition for perfect transfer) If the model $f=q\cdot\phi$ finds the invariant feature $z_{inv}$ and the target domain is a linear combination of the source domain  $D_t\in\{D_t:D_t=\sum_{i=1}^n\pi_iD_s^i,\sum_{i=1}^n\pi_i=1|D_s^i\in D_s\}$, we have the following proposition holds:
    
    \textit{Accuracy Transfer}: Let $ln(E(P(D|f)))$ represents the log-likelihood of model $f$ in dataset $D$, then we have:  
    \begin{align}
    	ln(E(P(D_t|f)))=ln(&E(P(D^i_s|f)))=ln(E(P(D_s|f))), \nonumber\\&
    	\forall i \in (1,2,...,n).
    \end{align}
    
    \textit{Discriminant Risk Transfer}: Similar to above conclusion, we have:     
    \begin{align}
    	\varepsilon_{D_t}^f=\varepsilon_{D_s^i}^f=\varepsilon_{D_s}^f, \forall i \in (1,2,...,n),
    	    \end{align}
    	\end{theorem}
  The above theory tells us that ideally optimizing the likelihood loss and Discriminant Risk in the source domain can reduce the corresponding loss on the target domain. In this case, the optimization result will be the most ideal.
  \begin{pf}

 For convenience, we refer to the log-likelihood as $loss1_{D_s^i}$ and the Discriminant Risk as $loss2_{D_s^i}$ on any subset.
  First, when the model captures $z_{inv}\perp Y$, then for $\forall D_s^i,D_s^j\in D_s$, we have:
   \begin{align}
   		loss1_{D_s^i}&=E_{Y\sim P_{D_s^i},Z\sim P^z_{D^i_s},\hat{b} \sim q(b)}(R(\hat{b}(Z),Y))\\&=E_{Y\sim P_{D_s^i},Z\sim P^{z,Y}_{D_s^i},\hat{b} \sim q(b)}(R(\hat{b}(Z),Y))\\&=E_{Y\sim P_{D_j^i},Z\sim P^{z,Y}_{D_s^j},\hat{b} \sim q(b)}(R(\hat{b}(Z),Y))\\&=loss1_{D^j_s}.
   \end{align}
   Noting that we consider the probability of occurrence of each $D_s^i$ to be equally likely, which means  $P(D_s^i|D_s)=P(D_s^j|D_s)$, then we have: $loss1_{D_s}=\frac{1}{N}\sum_{i=1}^Nloss1_{D^i_s}=loss_{D_s^i},\forall i \in(1,...,N)$.
   In the same way we are able to prove the loss of the target domain:
   \begin{align}
   		loss1_{D_t}&=E_{D_t,\hat{b} \sim q(b)}(R(\hat{b}(z),Y))\\&=\iint R(\hat{b}(z),Y)P_{D_t}^{Y,Z}dydz\\&=\iiint R(\hat{b}(z),Y)(\sum_{i=1}^N\pi_iP_{D_s^i}^{Y,Z})dydzdw\\&=\sum_{i=1}^N\pi_i\iiint R(\hat{b}(z),Y)P_{D_s^i}^{Y,Z}dydzdw\\&=\sum_{i=1}^N\pi_iE_{D^i_s,\hat{b} \sim q(b)}(R(\hat{b}(z),Y))\\&=\sum_{i=1}^N\pi_iE_{D_s,\hat{b} \sim q(b)}(R(\hat{b}(z),Y)\\&=loss1_{D_s},
   \end{align}
     Similarly, we can extrapolate the result of $loss_2$, noting that $\hat{y}$ is only determined by $z$, when the model capture the $z_{inv}$, we have:
     \begin{align}
     		loss2_{D_s}&=E_{D_s}(JS(P(\hat{y}|Y),P(\hat{Y}|Y)))\\&=\sum_{j=1}^K \iiint JS(P(\hat{y_j})|Y_j,e),P(\hat{Y_j}|Y_j,e)dP_edP_{D_s^e}^{Z,Y}\\&=\sum_{j=1}^K\iint JS(P(\hat{y_j})|Y_j),P(\hat{Y_j}|Y_j))dP^{Z_{inv},Y}_{D_s}.
     \end{align}
    Noting that $P^{Z_{inv},Y}_{D_s}=P^{Z_inv,Y}_{D_s^i},\forall i \in (1,...,N)$,
     then we can prove in the same way that: $loss2_{D_t}=loss2_{D_s^i}=loss2_{D_t}$.
      \end{pf}
\section{Experiments}
\begin{table*}[width=0.65\textwidth]
	\begin{center}
		\caption{Experiment result of three real-world datasets.}
		\label{expt}
		\setlength{\tabcolsep}{5mm}{
			\begin{tabular}{@{}lllll@{}}
				\bottomrule
				\specialrule{0em}{1.5pt}{1.5pt}
				\midrule
				Algorithm & \multicolumn{1}{c}{\textbf{PACS}} & \multicolumn{1}{c}{\textbf{VLCS}} & \textbf{Office-Home}    & \multicolumn{1}{c}{\textbf{Avg}} \\ \midrule
				\multicolumn{5}{c}{Model using domain labels}                                                                                                  \\ \midrule
				Mixup     & $78.70\pm 1.37$                    & $74.71\pm 1.2$                     & $60.31\pm 0.3$           & $71.2$                          \\
				CORAL     & $82.54\pm 0.5$                     & $74.00\pm 0.6$                       & $62.74\pm 0.2$           & $73.1$                           \\
				MMD       & $80.89\pm 1.6$                     & $74.19 \pm 0.9$                    & $58.42\pm 0.4$           & $71.1$                          \\
				IRM       & $75.95\pm 2.8$                     & $74.38 \pm 1.7$                    & $51.45\pm 0.4$          & $67.2$                           \\
				GDRO      & $80.38\pm 0.5$                     & $73.86 \pm 0.6$                    & $58.00 \pm 0.2$            & $70.8$                           \\
				MLDG      & $80.37\pm 0.3$                     & $74.47 \pm 1.2$                    & $58.85 \pm 0.6$          & $71.5$                           \\
				VREx      & $81.15\pm 0.3$                     & $74.40 \pm 1.7$                    & $59.11 \pm 0.3$          & $71.6$                           \\
				SagNet    & $81.03\pm 0.5$                    & $74.38 \pm 0.8$                    & $60.95 \pm 0.7$          & $72.1$                           \\
				Bayes-IRM & $81.16\pm 0.4$                     & $74.67 \pm 1.3$                    & $59.39 \pm 0.3$          & $71.7$                           \\
				Fish      & $80.34\pm 1.3$                     & $75.19 \pm 0.9$                    & $58.91 \pm 0.2$          & $71.4$                           \\
				Fishr     & $81.20\pm 0.9$                     & $75.42 \pm 0.4$                    & $59.10 \pm 1.1$          & $71.9$                           \\ \midrule
				\multicolumn{5}{c}{Model without domain labels}                                                                                                \\ \midrule
				ERM       & $80.51\pm 1.3$                     & $74.57 \pm 1.5$                    & $58.67 \pm 0.3$          & $71.2$                           \\
				ARM       & $79.28\pm 0.9$                     & $74.33 \pm 0.9$                    & $56.71 \pm 0.4$          & $69.9$                           \\
				RSC       & $80.00 \pm 0.3$                    & $74.25 \pm 1.2$                    & $62.97\pm 0.2$                      &               $72.3$                   \\
				SD        & $82.24 \pm 1.0$                    & $74.95 \pm 0.9$                    & \pmb{$63.20 \pm 0.3$} & $73.4$                           \\
				SelfReg   & $81.82 \pm 1.1$                    & $75.25 \pm 1$                      & $62.73 \pm 0.4$          & $73.3$                           \\
				EQRM  &$81.71 \pm 0.5$ & $74.74 \pm 0.5$ &$60.30 \pm 0.3$& $72.2$\\
				SAGM& $82.32\pm0.8$ & $74.80\pm0.8$ & $61.30\pm0.9$& $72.8$ \\
				DRM       & \pmb{$82.63 \pm 0.6$}           & \pmb{$76.12 \pm 0.9$}           & $63.03 \pm 0.3$          & \pmb{$73.9$}                  \\ 				\hline
				\specialrule{0em}{1.5pt}{1.5pt}
				\bottomrule
		\end{tabular}}
	\end{center}
\end{table*}
   \subsection{Real-world datasets}
    We use widely used benchmarks, PACS\cite{ref47}, VLCS\cite{ref46}, and Office-Home\cite{ref48} which contain real-world photos to evaluate the performance of DRM. All three datasets contain four source domains, which is not a large number of source domains, but the number of distributions of some of the data is enough to test the actual performance of our algorithm. Here is a brief description of the dataset:
    
    PACS: contains four sub-datasets, including art, cartoon, photo, sketch, with 9991 , 224x224x3 photos and 7 categories.
    
    VLCS: contains four sub-datasets, including Caltech101, LabelMe, SUN09, VOC2007, which contains 10729, 224x2 24x3 pictures and 5 categories.
    
    Office-Home: contains four subsets, including art, clipart, product, real, which contains 15588, 224x224x3 images and 65 categories.
    \paragraph{Baseline} To showcase the superiority of our algorithm that does not rely on domain labels, we will compare it with two categories of methods. The first category includes Mixup \cite{ref37}, CORAL \cite{ref8}, MMD \cite{ref56}, IRM \cite{ref3}, GDRO \cite{ref34}, MLDG \cite{ref11}, VREx \cite{ref19}, SagNet \cite{ref27}, Bayes-IRM \cite{ref23}, Fish \cite{ref37}, and Fishr \cite{ref31}, which use domain labels. The second category, like ours, does not require domain labels and includes ERM, ARM \cite{ref39}, RSC \cite{ref16}, SD\cite{ref29}, Selfreg \cite{ref18}, QRM\cite{ref61} and SAGM\cite{ref62}. 
    We will conduct experiments using Domainbed\cite{ref13} for all of the above methods to examine the goal of stable prediction from multiple perspectives. Comparing our approach with these methods will help demonstrate the stability and theoretical applicability of our approach.
    \paragraph{Experiment Setting} 
    
    In our experiments, we utilize a pre-trained ResNet-18 from ImageNet as the common backbone for all models. For models without specific requirements, a linear classification layer is added post the pre-trained model, followed by fine-tuning using the default Adam optimizer. Our model feature a final layer with a separate Bayesian linear classification layer, independently optimized via the SGD optimizer with a learning rate of 5e-5. The prior distribution used a Gaussian distribution with a mean of 0 and a variance of 10, performing three samplings during each batch to evaluate the partial distribution, while the rest of the model maintained consistency with others.
    
    To evaluate our model, we designate one domain dataset at a time as the test set, and divide the remaining datasets into training and validation sets at a ratio of (0.8, 0.2). The larger portion is used for training, and the smaller one forms a complete validation set. A consistent batch size of 32 is set for all models, and we conduct five experiments with different random number seeds for dataset partitioning.
    
    Our model emerge as the optimal choice after hyperparameter tuning on the PACS dataset. The selected hyperparameters are then uniformly applied across all three datasets, with the CDR penalty term set at $\alpha=5$, and a sliding update with $\beta$ set to 0.95. The Adam trainer with a learning rate of 5e-5 is consistently employed for all models. Default values from domainbed are adopted for any additional hyperparameters, selectively activating the L2 parametric penalty. Default hyperparameter values are retained for other models for consistency.
    
    Throughout five experiments with different random seeds, we select the model with the highest accuracy on the overall validation set, aligning with both domain-invariant characteristic assumptions and practical application considerations.
   
    \paragraph{Results} Our results demonstrate that the DRM achieves the highest accuracy on the first two datasets, while on the third Office-Home dataset, our method achieves an accuracy of 63.03\%, which is only marginally lower than the optimal accuracy of 63.2\%. In conclusion, our approach exhibits superior performance on average. Additionally, our method exhibits enhanced stability and robustness, as evidenced by its lower standard deviation. Furthermore, our experimental tuning process revealed that the DRM model is relatively robust to hyperparameters. Significantly reduced accuracy is only observed when overly large or small values of $\alpha$ and $\beta$ are chosen. These observations serve to emphasize the stability and strong generalization capabilities of our model. It is also worth noting that in our experiments, the majority of domain label-based methods failed to surpass the performance of label-free methods. This observation suggests that domain label requirements impose more stringent demands on the generalization ability, inducing instability in the model. It may also necessitate additional time for tuning optimal hyperparameters. However, doing so may consequently yield a lower expected accuracy across all potential data.
    \subsection{Ablation experiments}
    The experiments presented above have provided evidence of the reliability and superiority of our model. Nonetheless, a comprehensive assessment of each component validity remains outstanding. Therefore, we conduct further analysis of the constructed losses and the structural role to address this gap.
 
    \begin{table}
    	\centering
    	       		\caption{Ablation experiments of Bayesian linear classfier and CDR on PACS dataset.}
    	       		\label{tb2}    			
    	\resizebox{0.35\textwidth}{!}{
    		\begin{tabular}{llll}
    			\bottomrule
    			\specialrule{0em}{1.2pt}{1.2pt}
    			\hline
    			\multicolumn{1}{c}{$l_{ERM}$}   & \multicolumn{1}{c}{$l_{DRM}$}    & \multicolumn{1}{c}{$l_{VI}$} & \multicolumn{1}{c}{accuracy}        \\ \hline
    				\specialrule{0em}{0.5pt}{0.5pt}
    			\ding{52} &  &  & $80.5\pm 1.3$  \\
    			& \ding{52}& &$78.5\pm 0.7$  \\  
    			\ding{52} & \ding{52} &  & $80.6\pm 1.3$  \\
    			\ding{52} &  & \ding{52} & $81.4 \pm 1.3$ \\
    			 & \ding{52} & \ding{52} & $81.7\pm 0.4$  \\	 
    			\ding{52} & \ding{52} & \ding{52} & $82.6\pm 0.4$  \\ \hline
    			\specialrule{0em}{1.2pt}{1.2pt}
    			\bottomrule
    		\end{tabular}}
    \end{table}

   In our experiments, we use the resnet-18 as the main model and keep all the important settings consistent with the real-world dataset experiments we describe earlier. The only change is that we exclude certain losses or architectures. We train the models five times. Specifically, we use cross-entropy $l_{ERM}$ for the model extraction layer, applied CDR in the extraction layer of the model $l_{DRM}$, and replace the last layer with a Bayesian linear layer $l_{VI}$. It is important to note that our classification layer always uses cross-entropy for classification, no matter the situation. So, the top model represents ERM, the bottom model is our full DRM method, and the middle one is a different combination.
    
   Our findings, presented in Table \ref{tb2}, suggest that excluding the Bayesian linear layer causing a 2\% decrease in model accuracy and increasing instability. Additionally, the result of the second row indicates that our model requires ERM loss intervention to motivate the feature extractor to find invariant features while ensuring that the features themselves contribute to classification. When comparing the models using $l_{ERM}$ and $l_{DRM}$ separately and applying the Bayesian linear layer, we observe a decrease in accuracy to a similar level. The model using only $l_{ERM}$ exhibits a marginally higher accuracy but greater variance, while the model using only $l_{DRM}$ has slightly lower accuracy but significantly reduced variance. In summary, the Bayesian linear classification layer helps to further improve the generalization ability and stability of the model, and our loss function effectively guides the model towards discovering a more suitable model for generalized prediction, highlighting its efficacy.    
\section{Conclusion}
This paper presents a novel Theorem and approach for domain generalization without domain labels, which achieves state-of-the-art performance on multiple datasets. Our method aligns model prediction distributions on partial and overall data, and is compatible with existing methods. Furthermore, our approach can be used to promote fairness by maintaining consistency in model outputs. Overall, our work contributes to advancing the field of domain generalization and offers promising directions for future research.

\bibliographystyle{cas-model2-names}

\bibliography{Reference}

\end{document}